\pdfoutput=1

\documentclass[11pt]{article}

\usepackage{acl}

\usepackage{times}
\usepackage{latexsym}

\usepackage[T1]{fontenc}

\usepackage[utf8]{inputenc}

\usepackage{booktabs}
\usepackage{multirow}
\usepackage{tipa}
\usepackage{array}
\usepackage{footnote}
\usepackage{tablefootnote}
\usepackage{makecell}
\usepackage{url}
\usepackage{hyperref}
\usepackage{comment}
\usepackage{enumitem}
\usepackage{subcaption}
\usepackage{graphicx}
\usepackage{dirtytalk}
\usepackage{inconsolata}

\usepackage{microtype}

\title{An Investigation of Indian Native Language Phonemic Influences on L2 English Pronunciations}

\author{
Shelly Jain$^1$ \qquad Priyanshi Pal$^2$ \\
{ \bf Anil Vuppala$^1$ \qquad Prasanta Ghosh$^2$ \qquad Chiranjeevi Yarra$^1$ } \\
$^1$Speech Lab, Language Technologies Research Center (LTRC), IIIT, Hyderabad, 500032, India \\
$^2$Electrical Engineering, Indian Institute of Science (IISc), Bangalore, 560012, India \\ 
{\tt shelly.jain@research.iiit.ac.in}, {\tt \{priyanshipal, prasantg\}@iisc.ac.in} \\ \texttt{\{anil.vuppala, chiranjeevi.yarra\}@iiit.ac.in} }

\begin{document}

\maketitle


\begin{abstract}
Speech systems are sensitive to accent variations. This is especially challenging in the Indian context, with an abundance of languages but a dearth of linguistic studies characterising pronunciation variations. 
The growing number of L2 English speakers in India reinforces the need to study accents and L1-L2 interactions.
We investigate the accents of Indian English (IE) speakers and report in detail our observations, both specific and common to all regions. 
In particular, we observe the phonemic variations and phonotactics occurring in the speakers' native languages and apply this to their English pronunciations. 
We demonstrate the influence of 18 Indian languages on IE by comparing the native language pronunciations with IE pronunciations obtained jointly from existing literature studies and phonetically annotated speech of 80 speakers. 
Consequently, we are able to validate the intuitions of Indian language influences on IE pronunciations by justifying pronunciation rules from the perspective of Indian language phonology.
We obtain a comprehensive description in terms of universal and region-specific characteristics of IE, which facilitates accent conversion and adaptation of existing ASR and TTS systems to different Indian accents.
\end{abstract}

\section{Introduction}
\label{sec:intro}

India has approximately 1,652 languages and dialects from five language families -- Indo-Aryan, Dravidian, Austro-Asiatic, Tibeto-Burman and Tai-Kadai \cite{dryer2013world}.
Besides languages native to India, English is a lingua franca in education, law and administration \cite{kim2011corpus,ng2012syllable}, gaining importance as a second language.
However, existing speech systems are unable to handle the increasing English use, as people impose their native languages (L1) behaviours on their spoken English (L2).
These influences result in extensive diversity in the spoken English. 
On the other hand, Indian languages possess similar phonology due to shared region or language family.
The diversity of native languages influences English in each region, with the collection of spoken English varieties within India being referred to as \textit{Indian English} (IE).

Many measures have been attempted to handle the IE diversity.
\citet{sen2003pronunciation,Baby2021NonnativeEL} proposed models to automatically extract pronunciation rules mapping General American English to IE. \citet{ganji2019iitg,yarra2019indic,huang2020construction,jain2021iecps} described IE data and provided lexicons at different levels of phonetic variation. 
\citet{kumar2007building} used voice conversion to generate IE speech and discovered that the pronunciation model is vital for good performance.
Existing linguistic studies describing IE characteristics examined single varieties, failing to handle the variability. 
A thorough investigation involving Indian native language phonology is crucial. By observing L1 phonology, their influence on L2 English speech can be determined.
To address this, we investigate Indian native language influence on English, using a diverse IE corpus \cite{yarra2019indic,priyanshi2022ie}. 
By describing IE behaviour, one can handle unseen IE varieties, either adapting existing systems to specific varieties or handling unusual pronunciations due to L1 influence.
This also assists in tasks like accent recognition, mispronunciation detection and diagnosis (MDD), and native language recognition.
Consequently, this will enable the development of speech applications designed for the diverse varieties of IE. 

In this paper, we perform linguistic analysis on IE pronunciation by considering the speakers' native languages. 
We organise the behaviours of 18 Indian languages using phonemes and phonotactics, apply them to English, and categorise the pronunciations as generalised ("universal") or regional.
We use Indian language characteristics to validate IE pronunciation rules compiled jointly from data and prior linguistic studies. Hence, we derive linguistic rules to empirically verify intuitions about Indian language influence on IE speech from the perspective of both L1 and L2.
Our results can help develop speech systems robust to Indian variation. 
The universal characteristics are useful for custom lexicons and tasks like Indian accented ASR and TTS; regional characteristics provide detailed accent information for MDD and accent conversion.

The paper is organised as follows. The preliminary work which forms the basis for this study is detailed in Section \ref{sec:pre}. We describe our procedure and compile our observations about native language characteristics in Section \ref{sec:obs}. In Section \ref{sec:disc}, we discuss which Indian native language characteristics are realised in Indian English and which characteristics are verified by the study. Some applications are provided in Section \ref{sec:apps}. We conclude the paper in Section \ref{sec:conc}. Finally, we address some limitations of our approach in Section \ref{sec:lim}.

\section{Preliminary Study}
\label{sec:pre}

\subsection{Corpus}
\label{sec:data}

We consider Indic TIMIT, consisting of IE speech recordings of 80 speakers from different demographic regions with a few Indian states --  Northeast, East, North, Central, West and South \cite{yarra2019indic}. 
Each region is composed of a few Indian states. 
Since state boundaries were linguistically motivated, geographical and linguistic boundaries are aligned \cite{statesreorganizationact1956}. Hence, the native languages of the speakers from each of the regions are among the languages prominent in the respective region. 

They were divided into the following five groups, with and equal number of speakers: 

\begin{itemize} 
    \item \textbf{Group 1}: (Northeast and East) Maithili, Nepali, Oriya, Bengali, Assamese, Dimasa, Mog, Manipuri 
    \item \textbf{Group 2}: (North and Central) Malwi, Marwari, Hindi, Punjabi 
    \item \textbf{Group 3}: (West) Gujarati, Marathi, Konkani 
    \item \textbf{Group 4}: (Upper South) Kannada, Telugu 
    \item \textbf{Group 5}: (Lower South) Malayalam, Tamil 
\end{itemize}

Manipuri, Mog and Dimasa are Tibeto-Burman, Groups 4 and 5 are Dravidian, and the rest Indo-Aryan. 
Also, the Austro-Asiatic family influences Bengali, Assamese and Nepali, and the Indo-Aryan family influences Kannada and Telugu. Assamese is further influenced by the Tibeto-Burman family.

Indic TIMIT also consists of phonetic transcriptions for 2,342 recordings, annotated by two linguists. \citet{priyanshi2022ie} obtained annotations for 15,974  recordings, with over 190 recordings from native speakers (omitting Manipuri). In these recordings, the managed to cover the entire TIMIT stimulus spoken by native speakers from each region. Uniform influence from the region's languages was assumed. They reported good intra-rater agreement with Cohen's Kappa score \cite{cohen1960coefficient} of 0.827.

\begin{table*}[ht]
\centering
\begin{tabular}{cccccp{0.125\textwidth}<{\centering}c}
\toprule
\multirow{2}{*}{\textbf{Rule No.}} & \multicolumn{2}{c}{\textbf{Category 1}} & \multicolumn{2}{c}{\textbf{Category 2}} & \multicolumn{2}{c}{\textbf{Category 3}} \\
\cmidrule(lr){2-3}\cmidrule(lr){4-5}\cmidrule(lr){6-7}
& RP & IE & RP & IE & RP & IE \\
\midrule
1 & /\textipa{E}/ & /\textipa{e}/ & /\textipa{U}/ & /\textipa{u}/ & /\textipa{n}/ & /\textipa{@ n}/ \\
2 & /\textipa{\textturnv}/ & /\textipa{@}/ & /\textipa{aU}/ & /\textipa{au}/ & */\textipa{S}/ & /\textipa{s}/ \\
3 & /\textipa{d}/, /\textipa{t}/ & /\textipa{\:d}/, /\textipa{\:t}/ & /\textipa{j U}/ & /\textipa{u}/ & */\textipa{v}/ & /\textipa{b h}/ \\ 
4 & /\textipa{T}/ & /\textipa{\|[t h}/, /\textipa{\|[t}/ &  /\textipa{\textrhookrevepsilon}/ & /\textipa{@ r}/ & */\textipa{f}/ & /\textipa{p h}/ \\ 
5 & /\textipa{D}/ & /\textipa{\|[d}/ & /\textipa{A}/ & /\textipa{a r}/ & /\textipa{oU}/ & /\textipa{o:}/ \\ 
6 & /\textipa{l}/ & /\textipa{@ l}/ & /\textipa{I d}/ & /\textipa{e \:d}/ & /\textipa{eI}/ & /\textipa{e:}/ \\ 
7 & */\textipa{z}/ & /\textipa{s}/ & /\textipa{S n}/ & /\textipa{@ n}/ & /\textipa{6}/ & /\textipa{O:}/ \\ 
8 & */\textipa{I}/ & /\textipa{i}/ & /\textipa{@ n}/ & /\textipa{e n}/ & /\textipa{@U}/, /\textipa{E@}/, /\textipa{Ie}/, /\textipa{A:}/, /\textipa{O:}/ & - \\ 
\bottomrule
\end{tabular}
\caption{IE pronunciation rules relative to RP. `*' indicates native language specific rules}
\label{tab:rules}
\end{table*}

\subsection{Previous findings from the study}
\label{sec:pre_res}

A few works characterised IE pronunciation relative to Received Pronunciation (RP) \cite{wells1982accents,bansal1990pronunciation,bansal1994spoken,mesthrie2008introduction,sailaja2012indian}.
\citet{priyanshi2022ie} analysed the variations of IE against RP by validating pronunciation rules from literature against the rules obtained in a data-driven manner from the phonetic transcriptions of 15,974 recordings, and also reporting rules newly derived from the data.
The comparison was done at word level between canonical RP transcriptions and manually annotated IE transcriptions. 

A set of rules was obtained, indicating IE variations relative to RP. 
They were segregated into three categories based on where they were observed, as shown in Table \ref{tab:rules}. 
The phones marked in red were absent in the data and could not be observed.


\begin{itemize}
    \item \textbf{Category 1}: This consists of all the phonetic rules mentioned in literature which were also obtained from the data-driven approach that was considered \cite{priyanshi2022ie}. 
    \item \textbf{Category 2}: The rules in this category were obtained from the data-driven approach based on the threshold criteria, but were absent in existing literature.
    \item \textbf{Category 3}: This category consists of rules which occurred only in the literature, but were not obtained from the data-driven approach. The rules of this category were not obtained for either of two reasons. The first reason was the pronunciation rules produced by the data happened to be contradictory to rules regarding the same phones that were obtained from literature study. The second reason was that, while a set of rules in the literature described the effects on phones, the phones in these rules were absent from the data and hence could not be analysed.
\end{itemize}

Though \citet{priyanshi2022ie} considered a large number of phonetic transcriptions, the rules of Category 2 were absent from literature. This has two possible reasons: (1) existing literature failed to consider all IE varieties; or (2) the new data-driven rules were incorrect. To analyse this, we study the influences of Indian native languages (L1) on IE (L2) and validate the new rules. We also analyse how these influences could result in the rules reported in the existing literature as seen in Categories 1 and 3.

\section{Observations}
\label{sec:obs}

To analyse L1 influence on IE (L2), we study their properties. We profess that observing sounds and their changes in L1 provides insights into the pronunciation variations in L2. Hence, we study each Indian language individually and in relation to other Indian languages which are geographically or genealogically close. 
We study the L1 phonemes to understand the fundamental variations and categorise the phonemes based on their presence in Indian languages. The greater the number of languages having a phoneme, the greater the salience of that phoneme in the accent of L2.
We also study L1 phoneme interactions through phonotactics, which determine the valid phoneme sequences. Speakers tend to enforce L1 phonotactics on L2 during speech, so by observing these we can understand pronunciation variations in L2 due to L1.

Table \ref{tab:langs} details the 18 native Indian languages forming the subject our analysis, along with their corresponding groups, regions of origin, language family, and number of unique phonemes.

\begin{table*}[ht]
\centering
\begin{tabular}{cccccc}
\toprule
\textbf{No.} & \textbf{Group} & \textbf{Region} & \textbf{Language Family} & \textbf{Language} & \textbf{Phoneme Count} \\ 
\midrule
1 & \multirow{7}{*}{Group 1} & East & Tibeto-Burman &  Dimasa & 21 \\ 
2 & & East & Tibeto-Burman & Mog & 36 \\ 
6 & & East & Indo-Aryan & Maithili & 56 \\ 
5 & & East & Indo-Aryan & Oriya & 49 \\ 
4 & & East and Northeast & Indo-Aryan & Bengali & 47 \\ 
3 & & Northeast & Indo-Aryan & Assamese & 40 \\ 
7 & & Northeast & Indo-Aryan & Nepali & 46 \\ 
\midrule
8 & \multirow{4}{*}{Group 2} & North & Indo-Aryan & Punjabi & 51 \\ 
9 & & North & Indo-Aryan & Marwari & 50 \\ 
10 & & North and Central & Indo-Aryan & Hindi & 55 \\ 
11 & & Central & Indo-Aryan & Malwi & 46 \\ 
\midrule
12 & \multirow{3}{*}{Group 3} & West & Indo-Aryan & Gujarati & 47 \\ 
13 & & West & Indo-Aryan & Marathi & 48 \\ 
14 & & West & Indo-Aryan & Konkani & 55 \\
\midrule
15 & \multirow{2}{*}{Group 4} & Upper South & Dravidian &  Kannada & 50 \\ 
16 & & Upper South & Dravidian & Telugu & 55 \\
\midrule
17 & \multirow{2}{*}{Group 5} & Lower South & Dravidian & Malayalam & 54 \\
18 & & Lower South & Dravidian & Tamil & 37 \\
\bottomrule
\end{tabular}
\caption{Native Indian languages of the speakers}
\label{tab:langs}
\end{table*}

\subsection{Universal characteristics}
\label{sec:uni}

Most major Indian languages share phoneme inventories, including those in this study \cite{kishore2002data}. 
We call the shared properties universal characteristics. Some examples seen are the replacement of alveolar stops (/\textipa{t}/, /\textipa{d}/) with dental (/\textipa{\|[t}/, /\textipa{\|[d}/) and retroflex (/\textipa{\:t}/, /\textipa{\:d}/) stops, and the phonemic distinction of aspirated and unaspirated consonants. Indian languages also possess a greater number of vowels, though there is not much overlap with native English vowels.
Additionally, Indian languages have syllabic orthography with the inherent vowel schwa (/\textipa{@}/).

\begin{figure*}[ht]
    \centering
    \includegraphics[width=\textwidth]{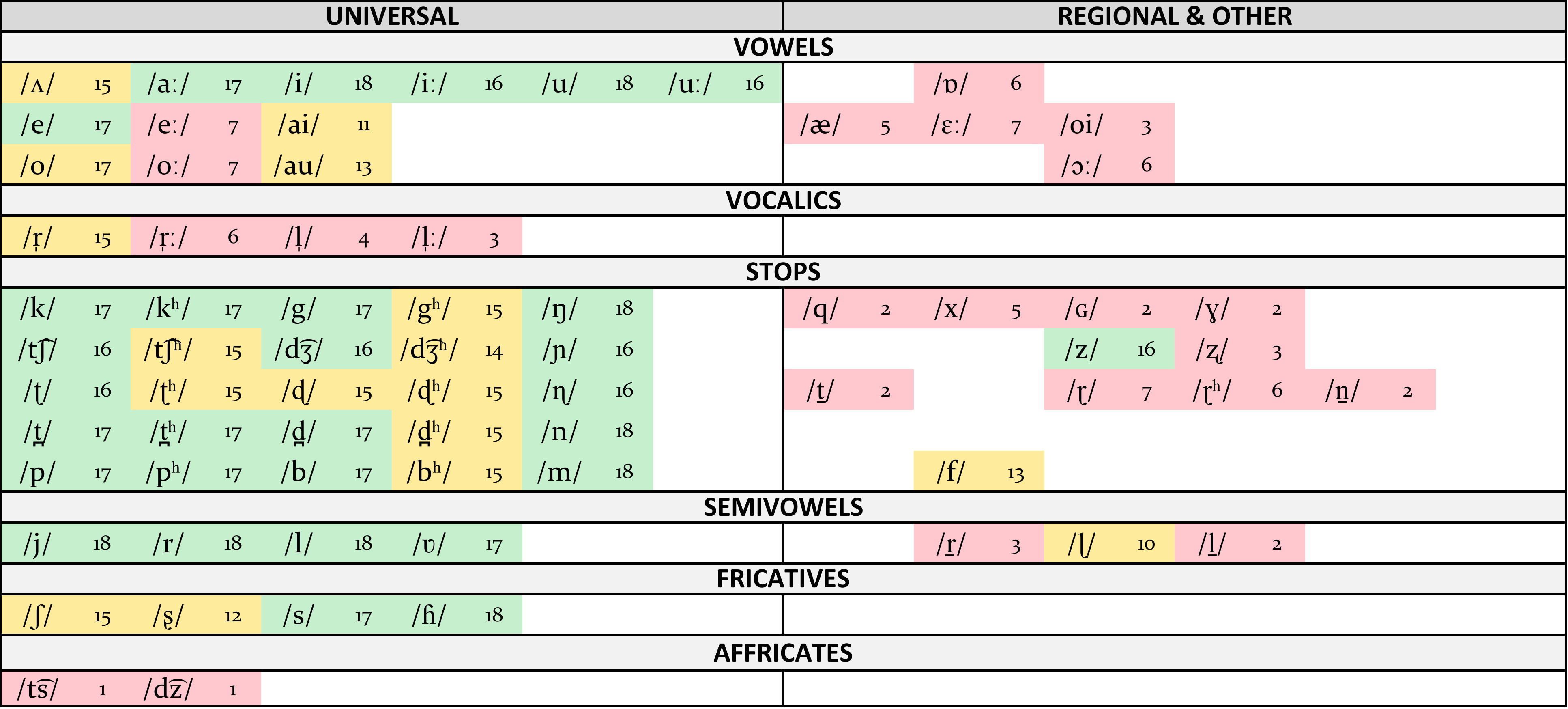}
    \caption{Frequency of phonemes in Indian languages (\textcolor{green}{green}: high, \textcolor{yellow}{yellow}: moderate, \textcolor{red}{red}: low)}
    \label{fig:phones}
\end{figure*}

Figure \ref{fig:phones} shows the frequency corresponding to the occurrence of phonemes in the 18 Indian languages studied. The phonemes are arranged in orthographic convention with vowels followed by consonants. The stops are arranged by place (velar, palatal, retroflex, dental, bilabial) and manner (unvoiced-unaspirated, unvoiced-aspirated, voiced-unaspirated, voiced-aspirated) of articulation\footnote{The second row of stops contains postalveolar affricates, but kept with the stops to keep consistent with the nasals.}. In the figure, the left side, \say{Universal}, refers to the common phonemes part of the phoneme inventories of most Indian languages. The right side, \say{Regional \& Other}, includes region-specific and foreign language phonemes. Borrowed phonemes (e.g. Perso-Arabic consonant /\textipa{q}/) and regional phonemes (e.g. Bengali vowel /\textipa{oi}/) are kept with associated standard phonemes (/\textipa{k}/ and /\textipa{ai}/ respectively). 

A total of 70 unique phonemes were observed in the given 18 languages. After recording the number of languages each phoneme was present in, we applied a percentile ranking to determine their degree of prevalence in Indian languages. This indicates whether the phoneme is universal, frequent or rare, by applying a uniform threshold on their frequencies. In this case, the top one-third ($\geq$ 66.7 percentile) phonemes occurred in 16 or more languages and the middle one-third ($\geq$ 33.3 percentile) phonemes occurred in 10 or more languages, while the bottom one-third phonemes occurred in fewer than 10 languages. These categories are indicated in red, yellow and green respectively. Of the 70 phonemes, 9 phonemes occurred in all 18 languages observed, of which it was surprising to note that only 2 were vowels.

Despite many similarities, Indian languages have specific features deviating from universal characteristics. These include tonality, phonemic stress, and vowel harmony. The greatest reason for deviation is difference in phoneme inventories -- best observed in Mog and Dimasa, the two Tibeto-Burman languages, with the English of their native speakers being distinct from that of other Indian languages.

\subsection{Regional characteristics}
\label{sec:belts}

We consider L2 English speech by L1 Indian language speakers uniformly from each region (Section \ref{sec:data}).
The languages of a region are expected to exhibit linguistic similarities due to either geography or genealogy. 
They do not generally deviate from the universal characteristics, but each region exhibits some unique features. Further, the behaviour of a few languages differs from the rest in the region. This could be due to genealogy and the external influence of neighbouring languages.

\subsubsection{Lower South}
\label{sec:belt_LS}

The prominent characteristics of these languages (Malayalam and Tamil) are vowel length distinction, and the lack of phonemic aspiration \cite{krishnamurti2003dravidian}. While Malayalam orthography has both aspirated and unaspirated consonants, actual pronunciation of the aspiration does not affect communication and is understood from context. On the other hand, Tamil lacks aspirated consonants entirely. Tamil also lacks voicing, unlike most Indian languages; voiced consonants can be identified based on the phonotactics of the language \cite{schiffman1999reference,keane2004illustrations}. In addition, both possess several consonants (mostly trills and flaps like /\textipa{\=*t}/, /\textipa{\=*n}/, /\textipa{\=*r}/, /\textipa{\=*l}/), which are not observed in other Indian languages \cite{hamann2003phonetics}.

\subsubsection{Upper South}
\label{sec:belt_US}

Like the languages of the Lower South, these (Telugu and Kannada) exhibit vowel length distinction and lack phonemic aspiration \cite{krishnamurti2003dravidian}. Both are similar to Malayalam, since aspirated consonants exist in their orthographies but are undifferentiated in speech. Word-initial vowels are also sometimes preceded by semivowels. For example, front vowels /\textipa{i}/ and /\textipa{e}/ are preceded by /\textipa{j}/, and back vowels /\textipa{u}/ and /\textipa{o}/ are preceded by /\textipa{V}/. Telugu, exhibits vowel harmony with /\textipa{u}/ \cite{wilkinson1974tense}, and also frequent voicing of medial consonants similar to Tamil \cite{bhaskararao2017telugu}.  

On comparing the Southern languages, it is clear that non-phonemic aspiration, vowel length distinction and consonant voicing are properties characteristic of Dravidian languages irrespective of region.

\subsubsection{West}
\label{sec:belt_W}

The Western languages, Konkani, Marathi and Gujarati, exhibit schwa deletion in a limited form, occurring word-finally in all, but also word-medially in Gujarati \cite{masica1993indo}. Konkani exhibits vowel length distinction (Marathi does so only in orthography), and is also the only language with a different inherent vowel /\textipa{\textbaro}/. Other unique properties are schwa fronting in Gujarati \cite{mistry1996gujarati}, lack of vowel nasalisation in Marathi, and consonant palatalisation in Konkani \cite{jain2007indo}.

\subsubsection{North}
\label{sec:belt_N}

Similar to Gujarati, these languages (Malwi, Marwari, Hindi and Punjabi) possess both vowel nasalisation as well as word-medial and word-final schwa deletion \cite{masica1993indo}. Hindi is more similar to Gujarati than the other Northern languages, as it exhibits schwa fronting \cite{shapiro2003hindi}. The commonality between them, observed across regions, can be attributed to close genealogical relation. Punjabi is distinct from the others due to tonality, gemination of consonants, and a unique form of vowel contrast. Punjabi vowels exist in three distinct groups (front, central, back), contrasted primarily by quality (central or peripheral) and secondarily by vowel length \cite{shackle2003panjabi}.

\subsubsection{East}
\label{sec:belt_E}

Unlike the other regions, the Eastern languages have three segments. The first, with Maithili and Nepali, most resembles the universal characteristics. In addition their vowels and semivowels are interchangeable \cite{khatiwada2009nepali,yadav2011reference}. The second segment, with Oriya, Bengali and Assamese, possesses a different inherent vowel /\textipa{O}/ \cite{masica1993indo}. Both the segments also exhibit limited schwa deletion and vowel nasalisation. The third segment, with Mog and Dimasa, differs considerably. The languages have reduced phoneme inventories compared to other regional languages, with much fewer vowels and consonants \cite{moral1997north,miri2003linguistic}. They also have different phonotactics with stricter sound rules. Thus, their phonemes and behaviour are quite distinct from other languages in the region. Due to geographical proximity to them, Assamese also bears some of their characteristics, especially the different consonants and vowel harmony \cite{mahanta2008directionality,mahanta2012assamese}.

\begin{table*}[hbt!]
\centering
\begin{tabular}{p{0.125\textwidth}<{\centering}p{0.25\textwidth}<{\centering}p{0.25\textwidth}<{\centering}p{0.25\textwidth}<{\centering}} 
\toprule
\multirow{2}{*}{\textbf{Language}} & \multicolumn{3}{c}{\textbf{Characteristics}} \\
\cmidrule{2-4}
& \textbf{Unique} & \textbf{Regional} & \textbf{Universal} \\
\midrule
\multirow{2}{*}{Tamil} & Lack of voiced consonants & Vowel length distinction & \multirow{26}{*}{Inherent vowel schwa /\textipa{@}/} \\
& Voicing of intervocalic consonants & Non-phonemic aspiration & \\
\cmidrule{1-2}
Malayalam & & Presence of trills and flaps & \\
\cmidrule{1-3}
\multirow{2}{*}{Telugu} & Vowel harmony & Vowel length distinction & \\
& Voicing of intervocalic consonants & Non-phonemic aspiration & \\
\cmidrule{1-2}
Kannada & Semivowels preceding word-initial vowels & \\
\cmidrule{1-3}
\multirow{3}{*}{Konkani} & Inherent vowel /\textipa{\textbaro}/ & Word-final schwa deletion & \\
& Consonant palatalisation & Vowel nasalisation & \\
& Vowel length distinction & & \\
\cmidrule{1-2}
Marathi & No vowel nasalisation & & \\
\cmidrule{1-2}
\multirow{2}{*}{Gujarati} & Word-medial schwa deletion & & \\
& Schwa fronting & & \\
\cmidrule{1-3}
Hindi & Schwa fronting & Schwa deletion & \\
\cmidrule{1-2}
\multirow{3}{*}{Punjabi} & Tonality & Vowel nasalisation & \\
& Consonant gemination & & \\
& Vowel contrasted by quality instead of length & & \\
\cmidrule{1-2}
Marwari & & & \\
\cmidrule{1-2}
Malwi & & & \\
\cmidrule{1-3}
Maithili & Complex phonotactics & Schwa deletion & \\
\cmidrule{1-2}
Nepali & & Vowel nasalisation & \\
& & Interchangeable vowels and semivowels & \\
\cmidrule{1-3}
Assamese & Vowel harmony & Schwa deletion & \\
\cmidrule{1-2}
Bengali & No consonant clusters & Vowel nasalisation & \\
\cmidrule{1-2}
Oriya & & Inherent vowel /\textipa{O}/ & \\
\cmidrule{1-3}
Mog & & Reduced phoneme set & \\
\cmidrule{1-2}
Dimasa & & & \\
\bottomrule
\end{tabular}
\caption{Universal, regional and unique characteristics of the 18 Indian languages}
\label{tab:il_characterictics}
\end{table*}

A comparison of West, North and East indicates that Indo-Aryan characteristics are schwa deletion and vowel nasalisation, with each branch showing unique properties. In contrast to this diversity, the Tibeto-Burman languages are characterised only by a reduced phoneme set. The influence of geographical proximity is evident in the similarities between Konkani and the Dravidian languages, and Assamese and the Tibeto-Burman languages.

All language characteristics have been compiled in Table \ref{tab:il_characterictics}. By observing the sound characteristics of each region, it is easy to see that genealogy and geography both considerably influence a language.

\section{Discussion}
\label{sec:disc}

We hypothesise that IE pronunciation variation (in Table \ref{tab:rules}) is due to L1 influence and exploit L1 behaviour to understand it. Since the pronunciation rules are derived from data that is unevenly distributed among the speakers of the 18 Indian languages, they are subject to statistical bias from languages with more speakers. By linking recorded IE behaviour to the characteristics observed in L1, we can attribute these characteristics to the number of associated languages, instead of number of speakers, which is balanced only by region. Consequently, we can determine whether these characteristics are universal or region-specific. Thus, the rules verified against L1 phonotactics will accurately describe the speakers' behaviour. 

\subsection{Pronunciation rules which were verified by L1 and L2 characteristics}

The first observation is consonant substitution. Indian languages lack alveolar stops and dental fricatives, so speakers substitute these with retroflex and dental stops, respectively. This is well accepted, corroborated by both data-driven and literature study of IE (Category 1, Rules 3, 4, 5), and examination of Indian languages (Section \ref{sec:uni}). Thus, these substitutions are universal IE characteristics. 

We also observe several regional characteristics. The prime example is interchangeable voiced and unvoiced consonants, from the data analysis (Category 1, Rule 7). Native language study allowed us to verify this as a behaviour of Dravidian language speakers, especially for Tamil or Telugu (Sections \ref{sec:belt_LS} and \ref{sec:belt_US}). This substitution is frequent due to a large speaker fraction; however, since the number of languages causing this is lower, the characteristic is not universal but regional to Upper and Lower South. Lack of aspiration is another regional characteristic, which we again observe in Dravidian language speakers -- the speakers substitute the unvoiced dental fricative /\textipa{T}/ with the unvoiced dental stop /\textipa{\|[t}/, while speakers of most other Indian languages use the aspirated unvoiced dental stop /\textipa{\|[t h}/ (Category 1, Rule 4).

The main vowel substitutions were the replacement of diphthongs with monophthongs. This is because, despite Indian languages having several diphthongs, few are shared with RP. Thus, they are approximated using known vowels. Prior literature predicted the resultant phones as long vowels, but the rules derived from data showed the substitutions as short vowels (Category 3, Rules 5, 6, 7). This seems surprising, but observing the native languages justifies the outcome. Of the 18 languages, only one-third distinguish vowel length, making the observed short vowels and predicted long vowels equivalent. Hence, for a universal characteristic, it is more important to correctly identify the substituting vowel --  usually, the diphthong's first vowel. 

The remaining vowel substitutions are as expected, where the vowel used has the most shared properties. For example, most Indian speakers tend to use tense vowels /\textipa{e}/, /\textipa{@}/, /\textipa{i}/ and /\textipa{u}/ in the place of corresponding lax vowels /\textipa{E}/, /\textipa{2}/, /\textipa{I}/ and /\textipa{U}/. This is seen in the preliminary study, in Table \ref{tab:rules} (Category 1, Rules 1, 2, 8; Category 2, Rules 1, 2). All vowel substitution rules are universal as they are validated by the universal characteristics of the native languages, regional behaviours being statistically insignificant.

\subsection{Discarded pronunciation rules}

Several predicted substitutions contradicted observed native language behaviours and were discarded. The first is palatalisation, i.e. the insertion of semivowel /\textipa{j}/ (Category 2, Rule 3). This is present in transcriptions but only occurs to a minor degree in the English words. Being almost non-existent in Indian languages, palatalisation is not distinguished from non-palatals. As a result, native Indian language speakers only perceive the distinct pronunciation of palatals. This direction requires future exploration for a better conclusion, hence we discard the substitution in this study. Similarly, the limited data and literature regarding gemination and the inconsistent occurrence of schwa insertion (Category 1, Rule 6; Category 2, Rule 7; Category 3, Rule 1) could not be validated from the perspective of L1-L2 interactions. Several rules were excluded due to lack of definite cause (Category 2, Rules 4, 5, 6, 8). 

Another kind of substitution discarded was the set of rules which could not be verified by data analysis, despite being observed in descriptions of both IE and native languages. In addition, we discarded certain behaviours predicted by the L1 phonotactics which were not observed in the data-driven study. The regional characteristics of Oriya, Bengali and Assamese (in Section \ref{sec:belt_E}), suggested the undifferentiated use of /\textipa{b}/, /\textipa{v}/, /\textipa{V}/, and /\textipa{w}/ in English (Category 3, Rule 3). Additionally, literature suggested that Telugu speakers tend to substitute /\textipa{S}/ with /\textipa{s}/ (Category 3, Rule 2), and that Gujarati and Marathi speakers tend to substitute /\textipa{f}/ with /\textipa{p h}/ (Category 3, Rule 4). However, contrary to expectation, L2 pronunciations in the preliminary study are consistent with those of other regions.

\section{Applications}
\label{sec:apps}

Modified lexicons are a popular method for dealing with non-native accents. A pronunciation dictionary for Indian English was devised by \citet{jain2021iecps}, to convert CMU dictionary \cite{dictionary1998carnegie} codes to the Common Phone Set (CPS) \cite{ramani2013common}, a set of phonetic codes which is popular in Indian multilingual speech systems. The dictionary substitutes the phones of General American English with the phones from native Indian languages, relying on the prominence of the substituted phones in Indian English to produce notable results. However, such a simple measure fails to deal with the variety in Indian English and is less effective when applied to specific regional accents. To resolve this issue, it is essential to have a larger dictionary with the sounds of all regional varieties. 

In addressing this aspect, the analysis in our work is beneficial. Since we provide descriptions of regional and general Indian English behaviours, it becomes simple to apply this knowledge and thus account for any pronunciation variations. The universal characteristics are useful for handling unseen accents of Indian English, and regional characteristics are useful for either adapting existing systems to specific Indian English varieties or handling unusual pronunciations due to native language influence.
By providing characteristics for the different regional varieties, identifying and thus correcting to and from desired varieties becomes possible.
In addition to accent adaptation, the Indian English profile can further assist in tasks like accent recognition, mispronunciation detection and diagnosis, and native language recognition by allowing the identification of region-specific behaviours.

\section{Conclusions}
\label{sec:conc}

In this paper, we study the phonemic variations and phonotactics of 18 Indian languages and analyse them against a data-driven study of IE pronunciations. From this, we compile the linguistic characteristics which cause universal and region-specific pronunciation variations in IE, validated from the perspective of both L1 and L2. Hence, we empirically verify intuitions about native language influence on Indian English.
Our linguistic profile is valuable for the development of speech systems robust to Indian context. 
The current analysis was on the phonemic space, but future efforts could explore the phonetic space by observing allophones in context.
Another avenue is the development of a phonotactic similarity index, which would allow for qualitative assessment of phonemic processes in languages and their influence on other languages.

\section{Limitations}
\label{sec:lim}

Certain conclusions drawn from the analysis are limited by the approach taken. The discussions detailed in Section \ref{sec:disc} rely upon the work done by \citet{priyanshi2022ie} and make use of their final list of rules describing IE pronunciation relative to RP. For this, we assume the data-driven study conducted by them, which contains 4 to 5 speakers per native language, is a sample which is representative of the full population of native speakers for each language. Additionally, without observing the data from Indic TIMIT, we cannot comment on the utterances themselves, or the effect of any possible phonological over-correction from the speakers while recording the sentences. 

\bibliography{anthology,custom}
\bibliographystyle{acl_natbib}

\end{document}